\documentclass{article}

 \PassOptionsToPackage{numbers, sort&compress}{natbib}

     \usepackage[preprint]{neurips_2020}

\usepackage[utf8]{inputenc} 
\usepackage[T1]{fontenc}    
\usepackage{hyperref}       
\usepackage{url}            
\usepackage{booktabs}       
\usepackage{amsfonts}       
\usepackage{nicefrac}       
\usepackage{microtype}      

\usepackage{xcolor}
\usepackage{amsmath}
\usepackage{bm}
\usepackage{graphicx}
\usepackage{subcaption}

\newcommand{\calW}{\mathcal{W}}
\newcommand{\tfea}{\theta_{\mbox{\tiny feature}}}
\newcommand{\ttra}{\theta_{\mbox{\tiny transform}}}

\newcommand{\codelink}{https://bit.ly/mil_pooling_filters}

\DeclareMathOperator*{\filter}{\theta_{\mbox{\tiny filter}}}

\title{Studying The Effect of MIL Pooling Filters\\on MIL Tasks}

\author{Mustafa Umit Oner$^{1,2}$, Jared Marc Song Kye-Jet$^{2}$, Hwee Kuan Lee$^{1,2,3,4}$ \& Wing-Kin Sung$^{1,5}$\\
	${^1}$School of Computing, National University of Singapore, Singapore, ${^2}$A*STAR Bioinformatics\\Institute, Singapore, ${^3}$Image and Pervasive Access Lab, CNRS UMI 2955, Singapore, ${^4}$Singapore\\Eye Research Institute, Singapore, ${^5}$A*STAR Genome Institute of Singapore, Singapore\\
	\texttt{\{umitoner,ksung\}@comp.nus.edu.sg}, \texttt{\{jskyejet\}@gmail.com}, \\\texttt{\{leehk\}@bii.a-star.edu.sg}
}

\begin{document}

	\maketitle

	\begin{abstract}
		There are different multiple instance learning (MIL) pooling filters used in MIL models. In this paper, we study the effect of different MIL pooling filters on the performance of MIL models in real world MIL tasks. We designed a neural network based MIL framework with 5 different MIL pooling filters: `max', `mean', `attention', `distribution' and `distribution with attention'. We also formulated 5 different MIL tasks on a real world lymph node metastases dataset. We found that the performance of our framework in a task is different for different filters. We also observed that the performances of the five pooling filters are also different from task to task. Hence, the selection of a correct MIL pooling filter for each MIL task is crucial for better performance. Furthermore, we noticed that models with `distribution' and `distribution with attention' pooling filters consistently perform well in almost all of the tasks. We attribute this phenomena to the amount of information captured by `distribution' based pooling filters. While point estimate based pooling filters, like `max' and `mean', produce point estimates of distributions, `distribution' based pooling filters capture the full information in distributions. Lastly, we compared the performance of our neural network model with `distribution' pooling filter with the performance of the best MIL methods in the literature on classical MIL datasets and our model outperformed the others.
	\end{abstract}
	
	\section{Introduction}
	Multiple instance learning (MIL) is a machine learning paradigm which learns the mapping between bags of instances and bag labels. In real life, MIL models are handy for tasks where data is in the form of bags of instances and only bag labels are provided. For example, medical image processing tasks are typical MIL tasks since an image can be treated as a bag of pixels (instances) and there usually exists only an image (bag) label without providing the particular region-of-interest~\cite{zhu2017deep}.
	
	Although \textit{positive vs negative bag classification}~\cite{dietterich1997solving, maron1998framework} is the most commonly used version of MIL tasks, there are other MIL tasks in the literature as well, like \textit{unique class count prediction}~\cite{Oner2020Weakly}, \textit{multi-class classification}~\cite{feng2017deep}, \textit{multi-task classification}~\cite{yang2016discriminative} or \textit{regression}~\cite{zhang2018bilateral}. In order to solve these MIL tasks, different methods have been developed. While some methods first classify instances inside bags and then pool the instances' scores by using an MIL pooling filter~\cite{dietterich1997solving,maron1998framework,andrews2003support,zhang2002dd}, others first extract features of instances inside bags, obtain bag level representations by using an MIL pooling filter and finally classify the bags~\cite{wang2018revisiting,ilse2018attention,Oner2020Weakly}. The common component in the two approaches is the MIL pooling filter. There are different MIL pooling filters, such as `max' pooling~\cite{wang2018revisiting,wu2015deep,feng2017deep}, `mean' pooling~\cite{wang2018revisiting,wang2019comparison} or `log-sum-exp' pooling~\cite{ramon2000multi}. However, their performance on different MIL tasks may be different. Indeed, it was, for example, shown that `linear softmax' pooling~\cite{dang2017deep} performed best among five different MIL pooling filters (`max', `mean', `linear softmax', `exponential softmax' and `attention') on MIL task of sound event detection~\cite{wang2019comparison}. Hence, the choice of MIL pooling filter based on MIL task is crucial.
	
	In this paper, our objective is to investigate the effect of different MIL pooling filters on the performance of an MIL model in a real world MIL task. With this objective, we designed neural network based MIL models with five different MIL pooling filters. We also formulated five different MIL tasks on a real world lymph node metastases dataset in~\cite{Oner2020Weakly} and applied our models on these tasks.
	
	The lymph node metastases dataset consists of images cropped from histopathology slides of lymph node sections~\cite{bejnordi2017diagnostic} and has corresponding ground truth metastases segmentation masks. There are three types of images in this dataset: \textit{fully normal} - all cells are normal, \textit{fully metastases} - all cells are metastases and \textit{boundary} - mixture of normal and metastases cells. We formulated five different MIL tasks on this dataset:  (i) \textit{positive vs negative bag classification}, (ii) \textit{unique class count prediction}, (iii) \textit{multi-class classification}, (iv) \textit{multi-task classification} and (v) \textit{regression}. In positive vs negative bag classification, our task is to predict whether an image contains metastases cells or not. Precisely, a positive bag is either a fully metastases image or a boundary image. A negative bag is a fully normal image. In regression, our task is to predict the percentage of metastases pixels inside the image. For the other MIL tasks, they are defined in Sec.~\ref{subsec:performance_analysis_of_mil_filters}. For each MIL task, we trained five different models with five different MIL pooling filters, namely (i) `max' pooling, (ii) `mean' pooling, (iii) `attention' pooling~\cite{ilse2018attention}, (iv) `distribution' pooling~\cite{Oner2020Weakly} and (v) `distribution with attention' pooling. We developed `distribution with attention' pooling by incorporating an attention mechanism similar to~\cite{ilse2018attention} into `distribution' pooling. Furthermore, we examined the power of exploiting the full information in distributions by using `distribution' pooling over using point estimates obtained by `mean' and `max' pooling in an MIL task of bag of metal balls classification. Lastly, although `distribution' pooling was used in a novel MIL task of unique class count prediction in~\cite{Oner2020Weakly}, it has never been tested on classical MIL datasets. Therefore, we tested its performance on classical MIL datasets and it outperformed state-of-the-art MIL methods. 
	
	Hence, there are four main contributions of this paper:
	\begin{enumerate}
		\item By examining the performance of MIL models with 5 different pooling filters on 5 different tasks formulated on the lymph node metastases dataset, we showed that different MIL pooling filters have different performances on different MIL tasks.
		\item We showed that `distribution' based pooling filters have better overall performance than point estimate based pooling filters, such as `mean' and `max' pooling since they capture full information in distributions rather than point estimates.
		\item We showed that our model with `distribution' pooling filter outperformed state-of-the-art MIL methods on classical MIL datasets.
		\item We developed a new MIL pooling filter, `distribution with attention' pooling filter, by incorporating an attention mechanism into `distribution' pooling filter.
	\end{enumerate}
	
	For the rest of the paper, Sec.~\ref{sec:mil_framework} introduces our MIL framework; Sec.~\ref{sec:related_work} gives a brief literature review; Sec.~\ref{sec:experiments} defines the experiments and presents the results and Sec.~\ref{sec:conclusion} concludes the paper.

	\section{Multiple Instance Learning Framework}\label{sec:mil_framework}
	In MIL paradigm, the objective is to predict a bag label $Y$ for a given bag of instances $X=\{x_i | x_i \in \mathcal{I}, i=1,2,\cdots,N\}$ where $ \mathcal{I}$ is the instance space and $N$ is the number of instances inside the bag. Each instance $x_i \in X$ is endowed with an underlying label $y_i \in \mathcal{L}$, where $\mathcal{L}$ is the instance label space; however, these labels are inaccessible during training. The relation between bag label $Y$ and instance labels ${\{y_i|y_i \in \mathcal{L}, i=1,2,\cdots,N\}}$ is imposed by the definition of MIL task. Note that although the number of instances in each bag can be different in real world, it is treated as constant in here for clarity of notation, yet all the properties stated in here are valid for bags with variable number of instances as well.
	
	\begin{figure}
		\centering
		\includegraphics[width=0.9\linewidth]{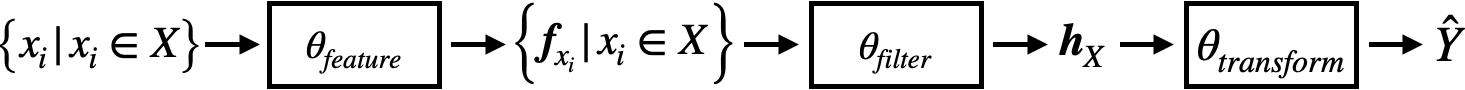}
		\caption{Block diagram of MIL framework. The \textit{feature extractor} module $\tfea$ extracts a feature vector $\bm{f}_{x_i} \in \mathcal{F}$, $\forall x_i \in X$. Then, the \textit{MIL pooling filter} module $\filter$ aggregates the extracted feature vectors and obtains a bag level representation $\bm{h}_{X} \in \mathcal{H}$. Lastly, the \textit{bag level representation transformation} module $\ttra$ transforms bag level representation into predicted bag label $\hat{Y} \in \mathcal{Y}$.}
		\label{fig:mil_block_diagram}
	\end{figure}
	
	Let $\mathcal{D}$ be an MIL dataset such that $\mathcal{D}=\{(X,Y) \ | \ X \in \mathcal{X} \ and \ Y \in \mathcal{Y}\}$, where $\mathcal{X} = \mathcal{I}^N$ is the bag space and $\mathcal{Y}$ is the bag label space. Given any pair $(X, Y) \in \mathcal{D}$, our objective is to predict bag label $Y$ for a given bag of instances $X=\{x_i | x_i \in \mathcal{I}, i=1,2,\cdots,N\}$. Let $\hat{Y}$ be predicted bag label of $X$. We obtain $\hat{Y}$ by using a three stage framework, block diagram of which is given in Figure~\ref{fig:mil_block_diagram}. The first stage is a \textit{feature extractor} module $\tfea:\mathcal{I} \rightarrow \mathcal{F}$, where $\mathcal{F}$ is the feature space. The \textit{feature extractor} module takes an instance as input and outputs a feature vector: $\bm{f}_{x_i} = \tfea(x_i) \in \mathcal{F}$ for each instance $x_i \in X$. The second stage is an \textit{MIL pooling filter} module $\filter:\mathcal{F}^{N} \rightarrow \mathcal{H}$, where $\mathcal{H}$ is the bag level representation space. The \textit{MIL pooling filter} module takes the set of extracted feature vectors $\bm{F}_{X}=\{\bm{f}_{x_i}| x_i \in X, i=1,\cdots N\}$ and aggregates them to obtain a bag level representation: $\bm{h}_{X}=\filter(\bm{F}_{X}) \in \mathcal{H}$. For example, `max' pooling gets maximum value of each feature over the feature vectors of all instances or `mean' pooling calculates the mean of each feature over the feature vectors of all instances. In this paper, we analyze the performance of 5 different MIL pooling filters (see Sec.~\ref{subsec:mil_pooling_filters}) on 5 different MIL tasks. The last stage is a \textit{bag level representation transformation} module $\ttra:\mathcal{H} \rightarrow \mathcal{Y}$. The \textit{bag level representation transformation} module transforms bag level representation into predicted bag label: $\hat{Y}=\ttra(\bm{h}_{X})$.
	
	\subsection{Modeling MIL with Neural Networks} \label{subsec:mil_with_nn}
	We use neural networks to implement the \textit{feature extractor} module $\tfea$ and the \textit{bag level representation transformation} module $\ttra$ so that we can fully parameterize the learning process. For \textit{MIL pooling filter} module $\filter$, we use different kinds of filters, some of which also incorporate trainable components parameterized by neural networks. This system of neural network modules are end-to-end trainable.
	
	Formally, given a bag of instances $X=\{x_i | x_i \in \mathcal{I}, i=1,2,\cdots,N\}$, for each $x_i \in X$, $\tfea$ module extracts $J$ features $[f_{x_i}^j | f_{x_i}^j \in \mathbb{R}, j=1,2, \cdots, J] =\bm{f}_{x_i} \in \mathcal{F}$ where $\mathcal{F}=\mathbb{R}^J$. After obtaining the set of extracted feature vectors $\bm{F}_{X}=\{\bm{f}_{x_i}| x_i \in X, i=1,\cdots N\}$, $\filter$ module obtains bag level representation $\bm{h}_{X} \in \mathcal{H}$ where $\mathcal{H}$ depends on $\filter$. Finally, $\ttra$ transforms $\bm{h}_{X}$ into predicted bag label $\hat{Y} \in \mathcal{Y}$. This notation is also used while defining MIL pooling filters in Sec.~\ref{subsec:mil_pooling_filters}.
	
	\subsection{MIL pooling filters}\label{subsec:mil_pooling_filters}
	An MIL pooling filter is used to aggregate instance level representations (extracted feature vectors of instances inside a bag) in order to obtain a bag level representation. As a key component of an MIL model, the MIL pooling filter provides the MIL model with two essential properties of MIL paradigm necessitated by set definition (note that a bag is nothing but a set). First, output of the model must be permutation-invariant, i.e. the model must produce the same output regardless of ordering of elements in the set. Second, the model must accept sets with variable sizes as input. These two properties are incorporated in our neural network model with the proper choice of the MIL pooling filters.

	\paragraph{Max pooling:}
	`Max' pooling filter obtains bag level representation by capturing maximum activation of each feature across the extracted features of instances inside a bag. 
	Bag level representation $\bm{h}_{X} = [h^j_{X} \ | \ h^j_{X} \in \mathbb{R}, j=1,2,\cdots,J] \in \mathcal{H}$ where $\mathcal{H}=\mathbb{R}^{J}$ and $h^j_{X} = \max_{i=1}^{N} f_{x_i}^{j} \ \forall_{j=1,2,\cdots,J}$.

	\paragraph{Mean pooling:}
	`Mean' pooling filter calculates bag level representation by getting the mean across the extracted features of all the instances inside a bag. 
	Bag level representation $\bm{h}_{X} = [h^j_{X} \ | \ h^j_{X} \in \mathbb{R}, j=1,2,\cdots,J] \in \mathcal{H}$ where $\mathcal{H}=\mathbb{R}^{J}$ and $h^j_{X} =\frac{1}{N} \sum_{i=1}^{N}f_{x_i}^{j} \ \forall_{j=1,2,\cdots,J}$.

	\paragraph{Attention pooling:}
	`Attention' pooling was introduced in~\cite{ilse2018attention}. In this method, bag level representation is obtained by weighted sum of extracted features of instances in a bag. Bag level representation $\bm{h}_{X} = [h^j_{X} \ | \ h^j_{X} \in \mathbb{R}, j=1,2,\cdots,J] \in \mathcal{H}$ where $\mathcal{H}=\mathbb{R}^{J}$ and $h^j_{X} = \sum_{i=1}^{N}w_if_{x_i}^{j} \ \forall_{j=1,2,\cdots,J}$. Note that each instance has an attention based weight obtained from a neural network module $\calW$ and instances' weights sum up to 1, as shown in Eq.~\ref{eq:instance_weight}. 
	
	
	\begin{equation}
	\label{eq:instance_weight}
	w_i = \frac{ \exp\{\calW( \bm{f}_{x_i} )\} }{\sum_{t=1}^{N} \exp\{\calW( \bm{f}_{x_t} ) \} }  \ \forall_{i=1,2,\cdots,N} 
	\end{equation}
	
	\paragraph{Distribution pooling:}
	`Distribution' pooling as MIL pooling filter was introduced in~\cite{Oner2020Weakly}. Firstly, marginal distribution of each extracted feature is estimated. Let $\tilde{p}^j_{X}(v)$ be the estimated marginal distribution of $j^{th}$ extracted feature. We obtain $\tilde{p}^j_{X}(v)$ by using kernel density estimation~\cite{parzen1962estimation}, which employs a Gaussian kernel with standard deviation $\sigma$, as shown in the Eq.~\ref{eq:kde}. Then, the estimated marginal distribution is sampled with $M$ bins as shown in Eq.~\ref{eq:kde_binning} to obtain corresponding $h^j_{X} \in \mathbb{R}^M$. Hence, bag level representation $\bm{h}_{X} = [h^j_{X} \ | \ h^j_{X} \in \mathbb{R}^M, j=1,2,\cdots,J] \in \mathcal{H}$ where $\mathcal{H}=\mathbb{R}^{MJ}$.
	
	\begin{equation}
	\label{eq:kde}
	\tilde{p}^j_{X}(v) = \frac{1}{N} \sum_{i=1}^{N}\frac{1}{\sqrt{2\pi{\sigma}^2}} e^{-\frac{1}{2{\sigma}^2} \left(v- f_{x_i}^{j}\right)^2} \ \forall_{j=1,2,\cdots,J}
	\end{equation}
	
	\begin{equation}
	\label{eq:kde_binning}
	h^j_{X} = \left[ \tilde{p}^j_{X}(v=v_b) \ | \ v_b=\frac{b}{M-1} \text{ and } b=0,1,\cdots,M-1 \right] \ \forall_{j=1,2,\cdots,J}  \text{ where } h^j_{X} \in \mathbb{R}^M
	\end{equation}

	Theoretically, full information in the extracted features can be captured by estimating the joint distribution; however, it is computationally intractable. Therefore, marginal distributions of extracted features are estimated to capture as much information as possible in `distribution' pooling. `Distribution' pooling has a notable advantage that it enables $\ttra$ module to fully utilize the information in shape of the distribution rather than looking at point estimates of it obtained by `mean' and `max' pooling filters. In principle, with a perfect $\filter$ and $\ttra$ modules, the point estimates obtained by `mean' and `max' pooling filters can be fully recovered from the estimated marginal distributions obtained by `distribution' pooling. 
	Hence, the information contained in the bag level representations derived by `max' and `mean' pooling filters are ``subsets'' of those derived by `distribution' pooling filter.
	
	\paragraph{Distribution with attention pooling:}
	We have extended `distribution' pooling as in Eq.~\ref{eq:kde_attention} by incorporating an attention mechanism similar to~\cite{ilse2018attention} in $\filter$ module. 
	While all of the instances are weighted equally (with a weight of $1/N$, see Eq.~\ref{eq:kde}) in `distribution' pooling, each instance $x_i \in X$, in this method, has an attention based weight $w_i$ obtained from a neural network module $\calW$ and instances' weights sum up to 1, as shown in Eq.~\ref{eq:instance_weight}. 
	After obtaining estimated marginal distribution of each extracted feature, we sample each estimated marginal distribution $\tilde{p}^j_{X}(v)\ \forall_{j=1,2,\cdots,J}$ with $M$ bins as in Eq.~\ref{eq:kde_binning} to obtain corresponding $h^j_{X} \in \mathbb{R}^M \ \forall_{j=1,2,\cdots,J}$. Hence, bag level representation $\bm{h}_{X} = [h^j_{X} \ | \ h^j_{X} \in \mathbb{R}^M, j=1,2,\cdots,J] \in \mathcal{H}$ where $\mathcal{H}=\mathbb{R}^{MJ}$. 
	
	\begin{equation}
	\label{eq:kde_attention}
	\tilde{p}^j_{X}(v) = \sum_{i=1}^{N} w_i \frac{1}{\sqrt{2\pi{\sigma}^2}} e^{-\frac{1}{2{\sigma}^2} \left(v- f_{x_i}^{j}\right)^2} \ \forall_{j=1,2,\cdots,J}
	\end{equation}
	
	\subsection{MIL tasks}  \label{subsec:mil_tasks}
	MIL paradigm in its most general form can actually be seen as a bag classification/regression problem. This generic problem can be formulated as different types of MIL tasks. In this section, we introduce five of such tasks used in our experiments. However, note that there are other MIL tasks as well~\cite{quadrianto2009estimating}.

	\paragraph{Positive vs negative bag classification:}
	This is the classical MIL task most commonly referred in the literature~\cite{dietterich1997solving, maron1998framework}. 
	In this task, each bag $X=\{x_i | x_i \in \mathcal{I}, i=1,2,\cdots,N\}$ has a bag label $Y = \max_{i=1}^{N}\{y_i\} \in \{0:negative, 1:positive\}$. 
	A bag is called `0:negative' if and only if all instances inside the bag are `0:negative', otherwise the bag is called `1:positive'. Note that each instance $x_i$ has a label $y_i \in \{0:negative, 1:positive\}$.
	
	\paragraph{Unique class count classification:}
	This is one of the recently introduced tasks in MIL~\cite{Oner2020Weakly}. 
	In this task, each bag $X=\{x_i | x_i \in \mathcal{I}, i=1,2,\cdots,N\}$ has a unique class count ($ucc$) label $Y = |\{y_i | y_i \in \{1, 2, \cdots, L\}, i=1,2,\cdots,N \}| \in \{1:ucc1, 2:ucc2, \cdots, L:uccL\}$. 
	\textit{Unique class count} is defined as the number of unique classes that all instances in the bag $X$ belong to.

	\paragraph{MIL multi-class classification:}
	Each bag $X=\{x_i | x_i \in \mathcal{I}, i=1,2,\cdots,N\}$ has a bag label $Y \in \{0,1\}^K$. 
	Label vector $Y$ is a one-hot vector consisting of $K$ classes.

	\paragraph{MIL multi-task classification:}
	Each bag $X=\{x_i | x_i \in \mathcal{I}, i=1,2,\cdots,N\}$ has a bag label $Y \in \{0,1\}^K$. 
	Label vector $Y$ is a binary vector consisting of $K$ classes, which may not be mutually exclusive, $Y = [Y^k | Y^k \in  \{0,1\}, k=1,2,\cdots,K]$. 

	\paragraph{Regression:}
	Each bag $X=\{x_i | x_i \in \mathcal{I}, i=1,2,\cdots,N\}$ has a bag level label $Y \in \mathbb{R}$.

	\section{Related work}\label{sec:related_work}
	MIL was first introduced in~\citep{dietterich1997solving,maron1998framework} as positive vs negative bag classification task for drug activity prediction. After that different versions of MIL tasks emerged: unique class count prediction~\cite{Oner2020Weakly}, multi-class classification~\cite{feng2017deep}, multi-task classification~\cite{yang2016discriminative} or regression~\cite{zhang2018bilateral}. In order to solve these MIL tasks, different types of MIL methods were derived with different assumptions~\citep{gartner2002multi, zhang2002dd, chen2006miles, foulds2008learning, zhang2009multi, zhou2009multi, ramon2000multi, zhou2002neural, zhang2004improve}, which are reviewed in detail in~\citep{foulds2010review}. However, recently there has been a massive shift towards to use neural networks in MIL setup to exploit the power of deep learning~\citep{wu2015deep, wang2018revisiting}.
	
	MIL methods were used for many different applications such as, image annotation / categorization / retrieval~\citep{chen2004image,zhang2002content,tang2010image}, text categorization~\citep{andrews2003support,settles2008multiple}, spam detection~\citep{jorgensen2008multiple}, medical image processing~\citep{dundar2007multiple, quellec2017multiple}, face/object detection~\citep{zhang2006multiple, felzenszwalb2010object}, object tracking~\citep{babenko2011robust}, defenses against adversarial attacks~\cite{Kou2020Enhancing}. Another related area is set classification, which is the same thing with bag classification in MIL since a bag is a set of instances. Recently, there are a few deep learning based studies in this area~\citep{zaheer2017deep,rezatofighi2017deepsetnet,kuncheva2017restricted}. In `Deep Sets'~\citep{zaheer2017deep}, for example, the general form of \textit{permutation invariant} set functions on a set is defined.
	
	Lastly, different types of MIL pooling filters are used to combine extracted features of instances inside the bags, such as `max' pooling~\cite{wang2018revisiting,wu2015deep,feng2017deep}, `mean' pooling~\cite{wang2018revisiting,wang2019comparison} or `log-sum-exp' pooling~\cite{ramon2000multi}. Although these filters have been widely used in the literature, recently there are new kinds of MIL pooling filters as well, such as `distribution' pooling~\cite{Oner2020Weakly}, `attention' pooling~\cite{ilse2018attention,lee2019set,wang2019comparison} or `sort' pooling~\cite{lu2015deep,Zhang2020FSPool}. We have also extended `distribution' pooling to `distribution with attention' pooling by incorporating an attention mechanism into it in this paper.
	
	\section{Experiments}\label{sec:experiments}
	We conducted three different types of experiments in this paper. (i) Performance analysis of different MIL pooling filters in different MIL tasks. (ii) Distribution pooling: exploiting the full information in distributions. (iii) Performance comparison on classical MIL datasets. 
	
	\subsection{Performance analysis of different MIL pooling filters in different MIL tasks}\label{subsec:performance_analysis_of_mil_filters}
	The objective of this experiment is to investigate the effect of MIL pooling filters on the performance of an MIL model in a real world MIL task. With this objective, we have designed a neural network based MIL framework with the same structure in Sec.~\ref{sec:mil_framework}. We used ResNet18~\cite{he2016deep} architecture without batch normalization as feature extractor module, $\tfea$, and a three layer multi-layer-perceptron as bag level representation transformation module, $\ttra$. We tested this framework with 5 different MIL pooling filters in $\filter$ module on 5 different MIL tasks formulated on the lymph node metastases dataset. Hence, we had 25 different models sharing the same architecture in $\tfea$ module. Note that the code was made publicly available at: \url{\codelink}

	\begin{table}
		\small
		\caption{MIL tasks defined on lymph node metastases dataset}
		\label{tab:mil_tasks}
		\centering
		\begin{tabular}{lccccc}
			\toprule
			& \multicolumn{5}{c}{bag label (Y)} \\
			\cmidrule(lr){2-6}
			& +ve/-ve & ucc  & 3-class & 2-task & \% metastases \\
			& (\textit{one-hot}) & (\textit{one-hot})  & (\textit{one-hot}) & (\textit{binary}) & (\textit{real-valued}) \\
			\cmidrule(lr){2-6}
			Fully normal & 10 & 10 & 100 & 1,0 & 0.0 \\
			Fully metastases & 01 & 10 & 010 & 0,1 & 1.0 \\
			Boundary & 01 & 01 & 001 & 1,1 & 0.0 < Y < 1.0 \\
			\midrule
			Loss & CCE & CCE & CCE & BCE & L1 \\
			\cmidrule(lr){1-6}
			\multicolumn{6}{l}{CCE: Categorical Cross Entropy~\cite{PyTorch_CategoricalCrossEntropy}, BCE: Binary Cross Entropy~\cite{PyTorch_BinaryCrossEntropy}} \\
			\bottomrule
		\end{tabular}
	\end{table}

	\begin{table}
		\small
		\caption{\textit{Top:} Test set performances of MIL models in 5 different MIL tasks formulated on the lymph node metastases dataset. \textit{Bottom:} Pairwise statistical test results are presented as color coded maps obtained by thresholding p-values at different significance levels. Best models are highlighted in bold. \textit{dist\_w\_att: `distribution with attention' pooling}}
		\label{tab:mil_results}
		\centering
		\begin{tabular}{lcccccc}
			\toprule
			& +ve/-ve & ucc  & 3-class &  \multicolumn{2}{c}{2-task (\textit{accuracy})} &  \% metastases\\
			\cmidrule(lr){5-6}
			& (\textit{accuracy}) & (\textit{accuracy})  & (\textit{accuracy}) & normal (N) & metastases (M) & (\textit{absolute error}) \\
			\cmidrule(lr){2-7}
			distribution & 0.8193 &  \textbf{0.8628} &  \textbf{0.7473} & 0.8533 & 0.8383 & 0.1906 \\
			dist\_w\_att & \textbf{0.9117} &  \textbf{0.8709} &  \textbf{0.7405} & 0.8696 &  \textbf{0.8723} &  \textbf{0.1671} \\
			mean & 0.8139 & 0.6413 & 0.6780 &  \textbf{0.8913} & 0.8438 & 0.2426 \\
			attention & 0.8804 & 0.6957 &  \textbf{0.7188} &  \textbf{0.8927} &  \textbf{0.8614} & 0.3264 \\
			max & 0.7636 & 0.7582 & 0.6712 & 0.8356 & 0.8111 & 0.2223 \\
			\midrule
			\multicolumn{7}{c}{ \includegraphics[width=0.96\linewidth]{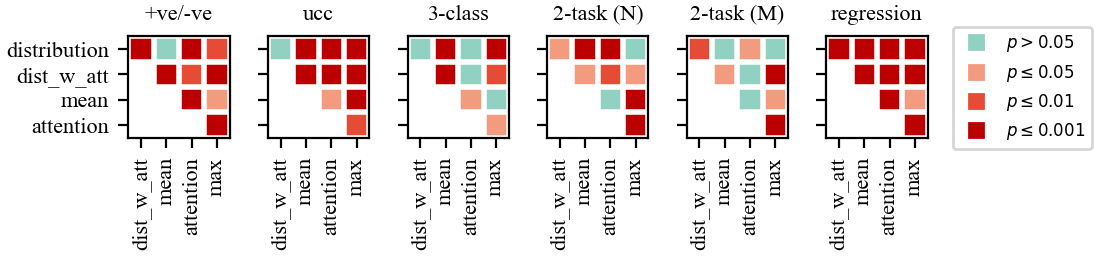} } \\ 
			\bottomrule
		\end{tabular}
	\end{table}

	The lymph node metastases dataset is adapted from~\cite{Oner2020Weakly} and has training, validation and test sets ({\textit{Supp.~1.1} for details}). The dataset consists of images cropped from histopathology slides of lymph node sections~\cite{bejnordi2017diagnostic} and has corresponding ground truth metastases segmentation masks. There are three types of images in this dataset: \textit{fully normal} - all cells are normal, \textit{fully metastases} - all cells are metastases and \textit{boundary} - mixture of normal and metastases cells. Similar to Sec.~\ref{subsec:mil_tasks}, we formulated five different MIL tasks on this dataset: positive vs negative bag classification  \textit{(+ve/-ve: predict whether an image contains metastases cells or not)}; unique class count prediction \textit{(ucc: predict how many types of cells exist in an image)}, multi-class classification \textit{(3-class: predict whether an image is fully normal, fully metastases or boundary)}, multi-task classification \textit{(2-task: $1^{st}$ task - predict whether an image contains normal cells or not, $2^{nd}$ task - predict whether an image contains metastases cells or not)} and regression \textit{(\% metastases: predict percentage of metastases pixels inside an image)}. The tasks are summarized by their bag labels for each kind of image and loss functions used during training in Table~\ref{tab:mil_tasks}.
		
	On each task, we trained five different models with MIL pooling filters defined in Sec.~\ref{subsec:mil_pooling_filters}, namely `max', `mean', `attention', `distribution' and `distribution with attention' pooling. Each model was randomly initialized and trained end-to-end with early-stopping criteria on validation set performance. Once we obtained the best models, we checked performances of the models on hold-out test set ({\textit{Supp.~1.2} for details}). The results are summarized in Table~\ref{tab:mil_results} together with used performance metrics for each task. Note that we presented the performances of two tasks separately for multi-task setup. Furthermore, we have conducted statistical tests for comparing the performance of different models in each task. For classification tasks, we used McNemar's test~\cite{everitt1977analysis} since all the models were trained on the same training set and tested on the same hold-out test set as suggested in~\cite{dietterich1998approximate}. On the other hand, we used paired t-test~\cite{hsu2005paired} on the absolute error values obtained for each sample in the test set to compare models in regression task. Results of the pairwise statistical tests for each task are also presented at the bottom of Table~\ref{tab:mil_results} with color coded maps obtained by thresholding p-values at different significance levels.
	
	We observed that while the performance of our framework in a particular MIL task is different for different MIL pooling filters, the performance of our framework with a specific MIL pooling filter also changes from task to task. Although there is no one single MIL pooling filter performing best for all of the MIL tasks, `distribution with attention' pooling filter seems to be a good candidate for almost all of the tasks except \textit{multi-task} setup, in which `attention' pooling seems to be a better candidate. Hence, the choice of MIL pooling filter based on MIL task is crucial for better performance. Please also note that `distribution' pooling outperforms `mean' and `max' pooling for almost all of the tasks.
	
	\subsection{Distribution pooling: exploiting the full information in distributions}\label{subsec:performance_on_metal_ball_classification}
\begin{figure}
		\centering
		\begin{subfigure}[b]{0.25\linewidth}
			\includegraphics[width=\linewidth]{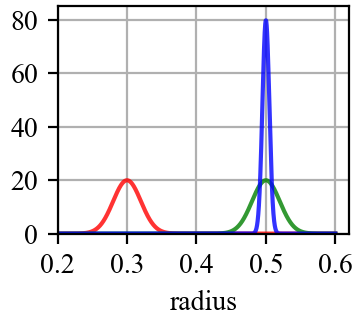}
			\caption{}
			\label{fig:r_distributions}
		\end{subfigure}
		\begin{subfigure}[b]{0.25\linewidth}
			\includegraphics[width=\linewidth]{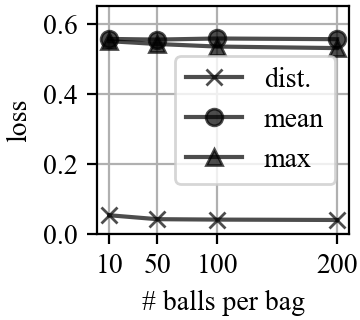}
			\caption{}
			\label{fig:loss_vs_num_balls_per_bag}
		\end{subfigure}
		\begin{subfigure}[b]{0.25\linewidth}
			\includegraphics[width=\linewidth]{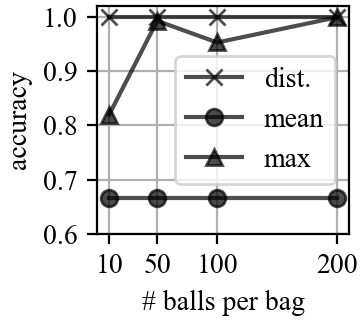}
			\caption{}
			\label{fig:acc_vs_num_balls_per_bag}
		\end{subfigure}
		\caption{3-class MIL classification task: classifying bags of metal balls from 3 different production lines. (a) Distribution of radius of balls produced in \textit{red}, \textit{green} and \textit{blue} production lines. (b) Loss and (c) accuracy values of MIL models with `distribution' (\textit{dist.}) pooling, `mean' pooling and `max' pooling filters on the test set bags with 10, 50, 100 and 200 balls per bag.}
		\label{fig:metal_balls}
	\end{figure}

	`Distribution' pooling extracts full information of the distribution. However, `mean' and `max' pooling only provide point estimates. This experiment aims to show that `distribution' pooling is more powerful than point estimators like `mean' and `max' pooling. Let's have a hypothetical factory producing metal balls for ball bearings. The factory has three production lines, namely \textit{red}, \textit{green} and \textit{blue}, and radius of the balls produced in these lines are normally distributed: $r_{red} \sim \mathcal{N}(\mu=0.3,\sigma=0.02)$, $r_{green} \sim \mathcal{N}(\mu=0.5,\sigma=0.02)$ and $r_{blue} \sim \mathcal{N}(\mu=0.5,\sigma=0.005)$, as shown in Figure~\ref{fig:r_distributions}. Then, our MIL task is to classify bags of metal balls from 3 different production lines, i.e. 3-class MIL classification task.
	
	We generated radius data for 900 bags of metal balls with corresponding production line labels. There were 300 bags from each production line and each bag had 200 balls. We designed an MIL framework such that a bag level representation was obtained by using an MIL pooling filter over the radius values of the balls in a bag. Then, the bag level representation was fed to a linear classifier to predict the production line label of the bag. We splitted data into training, validation and test sets with 600, 150 and 150 bags, respectively. In each set, there are equal number of bags from each class. We trained MIL models with `distribution', `mean' and `max' pooling filters on the bags with 10, 50, 100 and 200 balls per bag. All models were trained on the training set with categorical cross-entropy loss and fine-tuned on the validation set. Loss and accuracy values of the models on the test set are given in Figure~\ref{fig:loss_vs_num_balls_per_bag}~and~\ref{fig:acc_vs_num_balls_per_bag}, respectively. As it is seen in Figure~\ref{fig:acc_vs_num_balls_per_bag}, models with `distribution' pooling classified all the bags perfectly by exploiting the full information in the radius distribution of balls in the bags and outperformed the models with `mean' and `max' pooling. While models with `mean' pooling distinguished bags of \textit{red} line from others, it couldn't distinguish the bags of \textit{green} line from the bags of \textit{blue} line since their mean radius values were close to each other. For models with `max' pooling, they can distinguish bags of \textit{red} line from others. For bags from \textit{green} and \textit{blue} lines, the `max' polling can distinguish them only when the number of balls inside the bags is big enough. Moreover, loss of models with `distribution' pooling is much lower than the others, so they are much more confident in their predictions. {(See \textit{Supp.~2} for details.)}
	
	\subsection{Performance comparison on classical MIL datasets}\label{subsec:performance_on_classical_mil_datasets}
	This section compares the performance of our neural network model with `distribution' pooling filter (Distribution-Net) with the performance of the best MIL methods on classical MIL task of positive vs negative bag classification on 5 classical MIL datasets: drug activity prediction datasets MUSK1 and MUSK2~\cite{dietterich1997solving} and animal image annotation datasets FOX, TIGER and ELEPHANT~\cite{andrews2003support}. A drug is usually a small molecule that works by binding to a target protein molecule and may have multiple conformations. In MUSK datasets, each drug is represented as a bag of feature vectors obtained from its conformations and a label showing whether any of the conformations of the drug binds to its target protein (positive) or not (negative) is provided. Similarly, in animal datasets, each image is represented as a bag of feature vectors extracted from segments of the image and a label showing whether the image contains the particular animal (positive) or not (negative) is provided. {(See \textit{Supp.~3} for details of models and datasets.)}

	We used 10-fold cross validation and repeated each experiment 5 times. For each dataset, we have declared the mean of classification accuracies ($\pm$ standard error). We compared the performance of our neural network model with `distribution' pooling filter with the performance of state-of-the-art MIL methods on classical MIL datasets in Table~\ref{tab:classical_mil_datasets}. While first part of the table contains methods utilizing traditional machine learning techniques~\cite{andrews2003support,gartner2002multi,zhang2002dd,zhou2009multi,wei2016scalable}, second part of the table accommodates methods employing neural networks~\cite{wang2018revisiting,ilse2018attention}. The last part of the table shows the performance of our own model with `distribution' pooling filter (Distribution-Net), which outperformed all other methods on all datasets. While neural network based models generally outperformed the traditional machine learning based models, it seems that our Distribution-Net utilized the power of its `distribution' pooling filter over other neural network based models.
	
	 \setlength{\tabcolsep}{4pt}
	\begin{table}
		\small
		\caption{Performances of different MIL methods on classical MIL datasets. First part: methods utilizing traditional machine learning techniques. Second part: methods employing neural networks. Last part: our own model with `distribution' pooling filter (Distribution-Net).}
		\label{tab:classical_mil_datasets}
		\centering
		\begin{tabular}{llllll}
			\toprule
			METHOD & MUSK1 & MUSK2 & FOX & TIGER & ELEPHANT \\
			\midrule
			mi-SVM~\cite{andrews2003support} & 0.874 $\pm$ N/A & 0.836 $\pm$ N/A & 0.582 $\pm$ N/A & 0.784 $\pm$ N/A & 0.822 $\pm$ N/A \\
			MI-SVM~\cite{andrews2003support} & 0.779 $\pm$ N/A & 0.843 $\pm$ N/A & 0.578 $\pm$ N/A & 0.840 $\pm$ N/A & 0.843 $\pm$ N/A \\
			MI-Kernel~\cite{gartner2002multi} & 0.880 $\pm$ 0.031 & 0.893 $\pm$ 0.015 & 0.603 $\pm$ 0.028 & 0.842 $\pm$ 0.010 & 0.843 $\pm$ 0.016 \\
			EM-DD~\cite{zhang2002dd} & 0.849 $\pm$ 0.044 & 0.869 $\pm$ 0.048 & 0.609 $\pm$ 0.045 & 0.730 $\pm$ 0.043 & 0.771 $\pm$ 0.043 \\
			mi-Graph~\cite{zhou2009multi} & 0.889 $\pm$ 0.033 & 0.903 $\pm$ 0.039 & 0.620 $\pm$ 0.044 & 0.860 $\pm$ 0.037 & 0.869 $\pm$ 0.035 \\
			miVLAD~\cite{wei2016scalable} & 0.871 $\pm$ 0.043 & 0.872 $\pm$ 0.042 & 0.620 $\pm$ 0.044 & 0.811 $\pm$ 0.039 & 0.850 $\pm$ 0.036 \\
			miFV~\cite{wei2016scalable} & 0.909 $\pm$ 0.040 & 0.884 $\pm$ 0.042 & 0.621 $\pm$ 0.049 & 0.813 $\pm$ 0.037 & 0.852 $\pm$ 0.036 \\
			\midrule
			mi-Net~\cite{wang2018revisiting} & 0.889 $\pm$ 0.039 & 0.858 $\pm$ 0.049 & 0.613 $\pm$ 0.035 & 0.824 $\pm$ 0.034 & 0.858 $\pm$ 0.037 \\
			MI-Net~\cite{wang2018revisiting} & 0.887 $\pm$ 0.041 & 0.859 $\pm$ 0.046 & 0.622 $\pm$ 0.038 & 0.830 $\pm$ 0.032 & 0.862 $\pm$ 0.034 \\
			MI-Net with DS~\cite{wang2018revisiting} & 0.894 $\pm$ 0.042 & 0.874 $\pm$ 0.043 & 0.630 $\pm$ 0.037 & 0.845 $\pm$ 0.039 & 0.872 $\pm$ 0.032 \\
			MI-Net with RC~\cite{wang2018revisiting} & 0.898 $\pm$ 0.043 & 0.873 $\pm$ 0.044 & 0.619 $\pm$ 0.047 & 0.836 $\pm$ 0.037 & 0.857 $\pm$ 0.040 \\
			Attention~\cite{ilse2018attention} & 0.892 $\pm$ 0.040 & 0.858 $\pm$ 0.048 & 0.615 $\pm$ 0.043 & 0.839 $\pm$ 0.022 & 0.868 $\pm$ 0.022 \\
			Gated-Attention~\cite{ilse2018attention} & 0.900 $\pm$ 0.050 & 0.863 $\pm$ 0.042 & 0.603 $\pm$ 0.029 & 0.845 $\pm$ 0.018 & 0.857 $\pm$ 0.027 \\
			\midrule
			Distribution-Net (ours) & \textbf{0.923} $\pm$ 0.071 & \textbf{0.932} $\pm$ 0.067 & \textbf{0.680} $\pm$ 0.075 & \textbf{0.864} $\pm$ 0.054 & \textbf{0.900} $\pm$ 0.077 \\
			\bottomrule
		\end{tabular}
	\end{table}
		
	\section{Conclusion}\label{sec:conclusion}
	In this paper, we investigated the effect of different MIL pooling filters on the performance of an MIL model in some real world MIL tasks. We designed a neural network based MIL framework and analyzed the performance of our framework with 5 different MIL pooling filters in 5 different MIL tasks formulated on the lymph node metastases dataset. We observed that while the performance of our framework in a particular MIL task is different for different MIL pooling filters, the performance of our framework with a specific MIL pooling filter also changes from task to task. Thus, we conclude that choosing a correct MIL pooling filter for an MIL task is crucial for better performance. Furthermore, we noticed that although it is not the best performing MIL pooling filter for all tasks, `distribution with attention' pooling filter seems to be a good candidate for almost all MIL tasks. Actually, this is in accordance with our theoretical assertion in Sec.~\ref{subsec:mil_pooling_filters} that `distribution' based pooling filters capture richer information than point estimate based counterparts. 
	While our model with `distribution' pooling outperformed its counterparts with `mean' and `max' pooling in bag of metal balls classification experiment~(Sec.~\ref{subsec:performance_on_metal_ball_classification}), it also outperformed state-of-the art methods with different MIL pooling filters on classical MIL datasets~(Sec.~\ref{subsec:performance_on_classical_mil_datasets}). On top of observed superiority of `distribution' based pooling filters, in future, we want to explore them with fully trainable hyper-parameters, such as number of bins and standard deviation.
	
	\clearpage
	
	
	\begin{ack}
		This work is partly supported by the Biomedical Research Council of the Agency for Science, Technology, and Research, Singapore and the National University of Singapore, Singapore.
	\end{ack}
	
	\bibliographystyle{ieeetr}
	\bibliography{references}
	
	\clearpage

	\begin{center}
		\Huge
		\textbf{Supplementary Materials}
	\end{center}

	\section*{1 \hspace{.5em} Experiments on lymph node metastases dataset}
	We investigated the effect of MIL pooling filter on the performance of an MIL model in a particular real world MIL task. We designed a neural network based MIL framework and analyzed the performance of our framework with 5 different MIL pooling filters in 5 different MIL tasks formulated on a real world lymph node metastases dataset.
	
	Code for the experiments was made publicly available at: \url{\codelink}
	
	\subsection*{1.1 \hspace{.5em} Lymph node metastases dataset}
	The lymph node metastases dataset is adapted from~\cite{Oner2020Weakly}. The original dataset consists of $512\times512$ images cropped from histopathology slides of lymph node sections~\cite{bejnordi2017diagnostic} and has corresponding ground truth metastases segmentation masks. There are three types of images in this dataset: \textit{fully normal} - all cells are normal, \textit{fully metastases} - all cells are metastases and \textit{boundary} - mixture of normal and metastases cells. In order to make a clear distinction between these three types of images, we filtered out the images with: (i) $0< \text{percent metastases} \leq 20$ and (ii) $80 \leq \text{percent metastases} < 100$. Moreover, in order to obtain a balanced dataset, we dropped some of the images in the test set coming from one specific slide. The script to obtain our dataset from the original dataset in~\cite{Oner2020Weakly} is also inside the code folder.
	
\begin{figure}[ht]
		\centering
		\begin{subfigure}[b]{0.32\linewidth}
			\includegraphics[width=\linewidth]{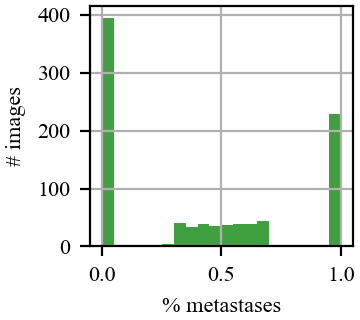}
			\caption{Training}
		\end{subfigure}
		\begin{subfigure}[b]{0.32\linewidth}
			\includegraphics[width=\linewidth]{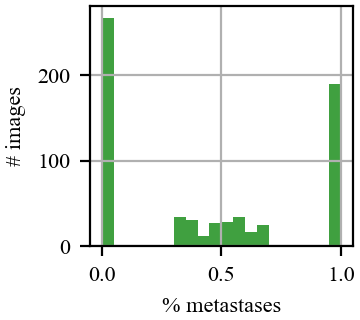}
			\caption{Validation}
		\end{subfigure}
		\begin{subfigure}[b]{0.32\linewidth}
			\includegraphics[width=\linewidth]{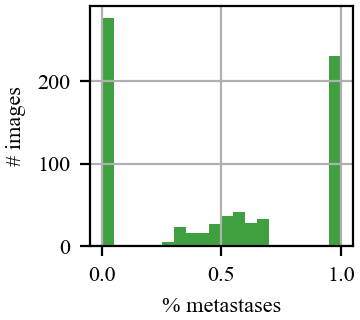}
			\caption{Test}
		\end{subfigure}
		\caption{Percent metastases histograms for training, validation and test sets.}
		\label{fig:percent_metastases_histograms}
	\end{figure}

	Our dataset is publicly available with this paper. For our dataset, number of images and percent metastases histograms in training, validation and test sets are given in Table~\ref{tab:number_of_images} and Figure~\ref{fig:percent_metastases_histograms}, respectively.
	
	\begin{table}[ht]
		\small
		\caption{Lymph node metastases dataset - number of images in training, validation and test sets.}
		\label{tab:number_of_images}
		\centering
		\begin{tabular}{lcccc}
			\toprule
			& Fully normal & Fully metastases & Boundary & Total \\
			\cmidrule(lr){2-5}
			Training & 395 & 228 & 310 & 933 \\
			Validation & 267 & 190 & 211 & 668 \\
			Test & 277 & 231 & 228 & 736 \\
			\bottomrule
		\end{tabular}
	\end{table}
	
	\subsection*{1.2 \hspace{.5em} Neural network architectures and hyper-parameters}
	We designed a neural network based MIL framework. We used ResNet18~\cite{he2016deep} architecture without batch normalization as feature extractor module, $\tfea$, and a three layer multi-layer-perceptron as bag level representation transformation module, $\ttra$. We tested this framework with 5 different MIL pooling filters in $\filter$ module on 5 different MIL tasks formulated. Hence, we had 25 different models sharing the same architecture in $\tfea$ module. Moreover, all models have the same hidden layers in $\ttra$ module; however, note that number of input nodes in $\ttra$ module depends on $\filter$ and number of output nodes in $\ttra$ module depends on MIL task. Please refer to the provided code for more details.
	
	During training of the models, each image was treated as a bag. We prepared bags on-the-go during training by randomly cropping $32\times32$ patches over the images. Each bag was created with 64 cropped patches (instances). Data augmentation was also applied on the cropped patches. We used batch size of 32 and extracted 32 features for each instance inside a bag. In the models with `distribution' pooling filter, kernel density estimation with a Gaussian kernel was used to estimate marginal distributions of extracted features. We used standard deviation of $\sigma=0.0167$ in Gaussian kernel and 21 bins to sample the estimated distributions. Furthermore, for attention mechanism, we used the same architecture in~\cite{ilse2018attention} in the models with `attention' pooling and `distribution with attention' pooling filters. We trained models by using ADAM optimizer with a learning rate of $lr=1e-4$ and $L2$ regularization on the weights with a weight decay of $weight\_decay=0.0005$. Each model was randomly initialized and trained end-to-end with early-stopping criteria on validation set performance. List of hyper-parameters used in MIL models is given in Table~\ref{tab:hyper_parameters}.
	
	During testing, we created 100 bags from each image and tested with the trained model. Final label for each image in the test set was obtained by averaging 100 predictions.
	
	\begin{table}[ht]
		\small
		\caption{Experiments on lymph node metastases dataset - architecture and list of hyper-parameters used in the MIL models. \textit{`dist\_w\_att': `distribution with attention' pooling} }
		\label{tab:hyper_parameters}
		\centering
		\begin{tabular}{lc}
			\toprule
			& input-$32x32x3$ \\
			& ResNet18 w/o BN \\
			& `distribution' /  `dist\_w\_att' / `mean' / `attention' / `max' pooling \\
			& Dropout(0.5) \\
			& fc-128 + ReLU \\
			Architecture & Dropout(0.5) \\
			& fc-32 + ReLU \\
			& Dropout(0.5) \\
			& fc-2 \textit{(+ve/-ve, ucc, 2-task)} / fc-3 \textit{(3-class)} / fc-1 \textit{(regression)} \\
			& softmax \textit{(+ve/-ve, ucc, 3-class)} / sigmoid \textit{(2-task)} / None \textit{(regression)} \\
			\cmidrule(lr){1-2}
			image size & $512\times512$ \\
			patch size & $32\times32$ \\
			\# instances per bag	& 64 \\
			\# features & 32 \\
			\# bins in `distribution' filters & 21 \\
			$\sigma$ in Gaussian kernel & 0.0167 \\
			Optimizer & ADAM \\
			Learning rate & $1e-4$ \\
			$L2$ regularization weight decay & 0.0005 \\
			batch size & 32 \\
			\bottomrule
		\end{tabular}
	\end{table}
	
	\section*{2 \hspace{.5em} Experiments on bags of metal balls}
	`Distribution' pooling extracts full information of the distribution. However, `mean' and `max' pooling only provide point estimates. This experiment aims to show that `distribution' pooling is more powerful than point estimators like `mean' and `max' pooling. We have a hypothetical factory producing metal balls for ball bearings. The factory has three production lines, namely \textit{red}, \textit{green} and \textit{blue}, and radius of the balls produced in these lines are normally distributed: $r_{red} \sim \mathcal{N}(\mu=0.3,\sigma=0.02)$, $r_{green} \sim \mathcal{N}(\mu=0.5,\sigma=0.02)$ and $r_{blue} \sim \mathcal{N}(\mu=0.5,\sigma=0.005)$. Then, our MIL task is to classify bags of metal balls from 3 different production lines, i.e. 3-class MIL classification task.
	
	We designed an MIL framework such that a bag level representation was obtained by using an MIL pooling filter over the radius values of the balls in a bag. Then, the bag level representation was fed to a linear classifier to predict the production line label of the bag. List of hyper-parameters used in MIL models is given in Table~\ref{tab:hyper_parameters_metal_balls}.
	
	\begin{table}[ht]
		\small
		\caption{Classifying bags of metal balls - architecture and list of hyper-parameters used in the MIL models.}
		\label{tab:hyper_parameters_metal_balls}
		\centering
		\begin{tabular}{lc}
			\toprule
			& input-1 \\
			Architecture & `mean' / `max' / `distribution' pooling \\
			& fc-3\\
			& softmax \\
			\cmidrule(lr){1-2}
			\# balls per bag	& 10 / 50 / 100 / 200 \\
			\# features & 1 \\
			\# bins in `distribution' pooling filters & 101 \\
			$\sigma$ in Gaussian kernel & 0.005 \\
			Optimizer & ADAM \\
			Learning rate & $1e-2$ \\
			$L2$ regularization weight decay & 0.0005 \\
			batch size & 64 \\
			\bottomrule
		\end{tabular}
	\end{table}
	
\begin{figure}
		\centering
		\includegraphics[width=0.9\linewidth]{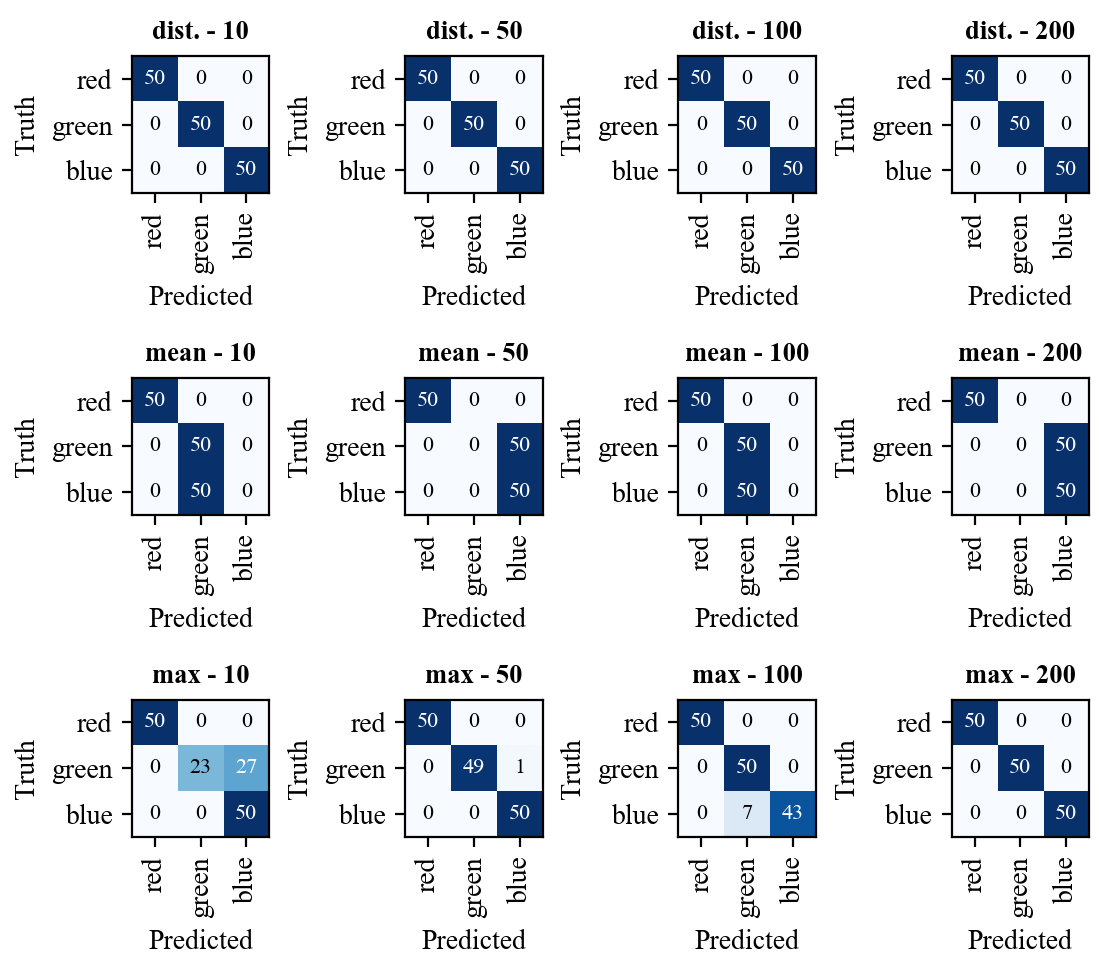}
		\caption{Confusion matrices for models with `distribution' \textit{(dist.)} pooling, `mean' pooling and `max' pooling filters on the test set bags with 10, 50, 100 and 200 balls per bag.}
		\label{fig:kdemeanmaxconfmat}
	\end{figure}

	Figure~\ref{fig:kdemeanmaxconfmat}, shows test set confusion matrices. Models with `distribution' pooling classified all the bags perfectly by exploiting the full information in the radius distribution of balls in the bags and outperformed the models with `mean' and `max' pooling. While models with `mean' pooling distinguished bags of \textit{red} line from others, it couldn't distinguish the bags of \textit{green} line from the bags of \textit{blue} line since their mean radius values were close to each other. For models with `max' pooling, they can distinguish bags of \textit{red} line from others. For bags from \textit{green} and \textit{blue} lines, the `max' polling can distinguish them only when the number of balls inside the bags is big enough.
	
	\newpage
	\section*{3 \hspace{.5em} Experiments on classical MIL datasets}
	This section compares the performance of our neural network model with `distribution' pooling filter (Distribution-Net) with the performance of the best MIL methods on classical MIL task of positive vs negative bag classification on 5 classical MIL datasets: drug activity prediction datasets MUSK1 and MUSK2~\cite{dietterich1997solving} and animal image annotation datasets FOX, TIGER and ELEPHANT~\cite{andrews2003support}. Summary of classical MIL datasets is given in Table~\ref{tab:summary_of_classical_mil_datasets}.
	
	\begin{table}[ht]
		\small
		\caption{Summary of classical MIL datasets}
		\label{tab:summary_of_classical_mil_datasets}
		\centering
		\begin{tabular}{lccccccc}
			\toprule
			& \multicolumn{3}{c}{\# bags} & \multicolumn{3}{c}{\# instances per bag} & \\
			\cmidrule(lr){2-4} \cmidrule(lr){5-7}
			& positive & negative & total & min & max & average & \# features \\
			\cmidrule(lr){2-8}
			MUSK1 & 47 & 45 & 92 & 2 & 40 & 5.17 & 166 \\
			MUSK2 & 39 & 63 & 102 & 1 & 1044 & 64.69 & 166 \\
			FOX & 100 & 100 & 200 & 2 & 13 & 6.6 & 230 \\
			TIGER & 100 & 100 & 200 & 1 & 13 & 6.1 & 230 \\
			ELEPHANT & 100 & 100 & 200 & 2 & 13 & 6.96 & 230 \\
			\bottomrule
		\end{tabular}
	\end{table}
	
	The summary of architectures and hyper-parameters used in MIL models on `MUSK' and `Animal' datasets are given in Table~\ref{tab:musk_datasets_summary_of_mil_models} and Table~\ref{tab:animal_datasets_summary_of_mil_models}, respectively. Note that we used mini-batch training with bags including equal number of instances. We created bags by sampling from available instances of each sample (a drug with multiple conformations for `MUSK' datasets and an image with multiple segments in `Animal' datasets). When the number of available instances of a sample is less than number of instances required to create a bag we used available instances more than once in a bag.
	
	\begin{table}[ht]
		\small
		\caption{MUSK datasets - architecture and list of hyper-parameters used in the MIL models.}
		\label{tab:musk_datasets_summary_of_mil_models}
		\centering
		\begin{tabular}{lc}
			\toprule
			& input-166 \\
			& fc-64 + ReLU \\
			& Dropout(0.5) \\
			& fc-32 + ReLU \\
			& Dropout(0.5) \\
			& fc-32 + Sigmoid \\
			Architecture & `distribution' pooling \\
			& Dropout(0.5) \\
			& fc-64 + ReLU \\
			& Dropout(0.5) \\
			& fc-32 + ReLU \\
			& Dropout(0.5) \\
			& fc-2 \\
			& softmax \\
			\cmidrule(lr){1-2}
			\# instances per bag & 16 \\
			\# features & 32 \\
			\# bins in `distribution' pooling filters & 11 \\
			$\sigma$ in Gaussian kernel & 0.1 \\
			Optimizer & ADAM \\
			Learning rate & $5e-4$ \\
			$L2$ regularization weight decay & 0.1 \\
			batch size & 8 \\
			\bottomrule
		\end{tabular}
	\end{table}
	
	\begin{table}[ht]
		\small
		\caption{Animal datasets - architecture and list of hyper-parameters used in the MIL models.}
		\label{tab:animal_datasets_summary_of_mil_models}
		\centering
		\begin{tabular}{lc}
			\toprule
			& input-230 \\
			& fc-256 + ReLU \\
			& Dropout(0.5) \\
			& fc-128 + ReLU \\
			& Dropout(0.5) \\
			& fc-64 + ReLU \\
			& Dropout(0.5) \\
			& fc-32 + Sigmoid \\
			Architecture & `distribution' pooling \\
			& Dropout(0.5) \\
			& fc-384 + ReLU \\
			& Dropout(0.5) \\
			& fc-192 + ReLU \\
			& Dropout(0.5) \\
			& fc-2 \\
			& softmax \\
			\cmidrule(lr){1-2}
			\# instances per bag & 16 \\
			\# features & 32 \\
			\# bins in `distribution' pooling filters & 11 \\
			$\sigma$ in Gaussian kernel & 0.1 \\
			Optimizer & Adam \\
			Learning rate & $5e-6$ \\
			$L2$ regularization weight decay & 0.1 \\
			batch size & 8 \\
			\bottomrule
		\end{tabular}
	\end{table}

\end{document}